\documentclass[conf]{new-aiaa}
\usepackage[utf8]{inputenc}
\usepackage{wrapfig}
\usepackage{booktabs}
\usepackage{graphicx}
\usepackage{amsmath}
\usepackage[version=4]{mhchem}
\usepackage{siunitx}
\usepackage{longtable,tabularx}
\usepackage{graphicx}
\title{Hall effect thruster design via deep neural network for additive manufacturing}

\author{Korolev K.V \footnote{Founder, Pure EP}}
\affil{Student, Moscow, Russia, korolev.konstantin.v@gmail.com}

\begin{document}

\maketitle

\begin{abstract}
Hall effect thrusters are one of the most versatile and popular electric propulsion systems for space use. Industry trends towards interplanetary missions arise advances in design development of such propulsion systems. It is understood that correct sizing of discharge channel in Hall effect thruster impact performance greatly. Since the complete physics model of such propulsion system is not yet optimized for fast computations and design iterations, most thrusters are being designed using so-called scaling laws. But this work focuses on rather novel approach, which is outlined less frequently than ordinary scaling design approach in literature. Using deep machine learning it is possible to create predictive performance model, which can be used to effortlessly get design of required hall thruster with required characteristics using way less computing power than design from scratch and way more flexible than usual scaling approach. 
\end{abstract}

\section{Nomenclature}

{\renewcommand\arraystretch{1.0}
\noindent\begin{longtable*}{@{}l @{\quad=\quad} l@{}}
$U_d$  & discharge voltage \\
$P$ &    discharge power \\
$T$& thrust \\
$\dot{m}_a$ & mass flow rate \\
$I_{sp}$ & specific impulse \\
$\eta_m$ &  mass utilization efficiency \\
$\eta_a$ &  anode efficiency \\
$j$ & $P/v$ [power density] \\
$v$ & discharge channel volume \\
$h, d, L$   & generic geometry parameters \\
$C_*$  & set of scaling coefficients \\
$g$  & free-fall acceleration \\
$M$  & ion mass
\end{longtable*}}

\section{Introduction}
\lettrine{T}{he} application of deep learning is extremely diverse, but in this study it focuses on case of hall effect thruster design. Hall effect thruster (HET) is rather simple DC plasma acceleration device, but due to complex and non linear process physics we don't have any full analytical performance models yet. Though there are a lot of ways these systems are designed in industry with great efficiencies, but in cost of multi-million research budgets and time. This problem might be solved using neural network design approach and few hardware iteration tweaks\cite{plyashkov_scaling_2022}.

Scaled thrusters tend to have good performance but this approach isn't that flexible for numerous reasons: first and foremost, due to large deviations in all of the initial experimental values accuracy can be not that good, secondly, it is hardly possible to design thruster with different power density or $I_{sp}$ efficiently. 

On the other hand, the neural network design approach has accuracy advantage only on domain of the dataset\cite{plyashkov_scaling_2022}, this limitations is easily compensated by ability to create relations between multiple discharge and geometry parameters at once. Hence this novel approach and scaling relations together could be an ultimate endgame design tool for HET.

Note that neither of these models do not include cathode efficiencies and performances. So as the neutral gas thrust components. Most correlations in previous literature were made using assumption or physics laws\cite{shagayda_hall-thruster_2013}, in this paper the new method based on feature generation, GAN dataset augmentation and ML feature selection is suggested.
\subsection{Dataset enlargement using GAN}
As we already have discussed, the data which is available is not enough for training NN or most ML algorithms, so I suggest using Generative Adversarial Network to generate more similar points. Generative model trains two different models - generator and discriminator. Generator learns how to generate new points which are classified by discriminator as similar to real dataset. Of course it is very understandable that model needs to be precise enough not to overfit on data or create new unknown correlations. Model was checked via Mean Absolute Percentage Error (MAPE) and physical boundary conditions. After assembling most promising architecture, the model was able to generate fake points with MAPE of $~4.7\%$. We need to measure MAPE to be sure point lie on same domain as original dataset, as in this work we are interested in sub-kilowatt thrusters. After model generated new points they were check to fit in physical boundaries of scaled values (for example thrust couldn't be more than 2, efficiency more than 1.4 and so on, data was scaled on original dataset to retain quality), only 0.02\% of points were found to be outliers. The GAN architecture and dataset sample is provided as follows.

\begin{figure}
    \begin{minipage}[c]{0.22\linewidth} 
        \centering
        \includegraphics[width=\textwidth]{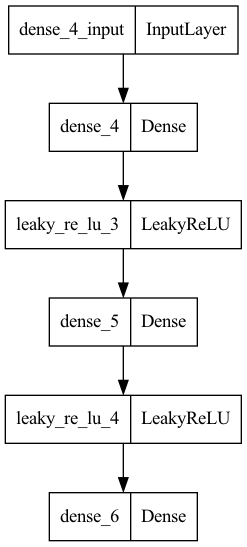}
        \caption{GAN generator architecture}
        \end{minipage}\hfill
    \begin{minipage}[c]{0.5\linewidth}
        \centering 
        \includegraphics[width=\textwidth]{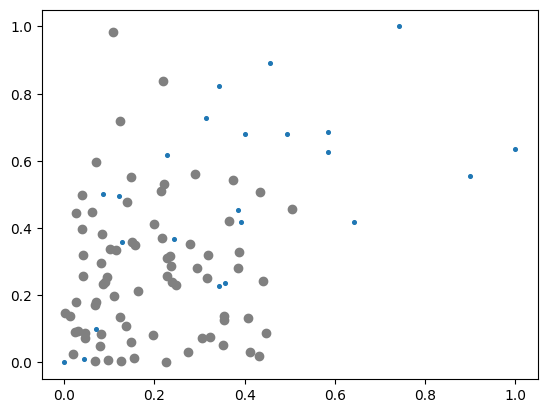}
        \caption{Sample of generated data ($d$,$T$), gray - fake, blue real}
        \end{minipage}
\end{figure}

\section{General Relations}

As we will use dataset of only low power hall thrusters, we can just ignore derivation of any non-linear equations and relations and use traditional approach here. Let's define some parameters of anode: 
\begin{equation}
\label{sample:equation}
\alpha = \frac{\dot{m}\beta}{{\dot{m}_a}}, 
\end{equation}
Where $\alpha$ is anode parameter of $\beta$ thruster parameter. This is selected because this way cathode and other losses wont be included in the model.
One of key differences in this approach is fitting only best and most appropriate data, thus we will eliminate some variance in scaling laws. Though due to machine learning methods, we would need a lot of information which is simply not available in those volumes. So some simplifications and assumptions could be made. Firstly, as it was already said, we don't include neutralizer efficiency in the model. Secondly, the model would be correct on very specific domain, defined by dataset, many parameters like anode power and $I_{sp}$ still are using semi-empirical modelling approach.
The results we are looking for are outputs of machine learning algorithm: specific impulse, thrust, efficiency, optimal mass flow rate, power density. Function of input is solely dependant on power and voltage range.
For the matter of topic let's introduce semi-empirical equations which are used for scaling current thrusters.

\begin{align}
&h=C_hd\\
&\dot{m_a} = C_m hd\\
&P_d=C_pU_dd^2\\
&T=C_t\dot{m_a}\sqrt{U_d}\\
&I_{spa}=\frac{T}{\dot{m_a} g}\\
&\eta_a=\frac{T}{2\dot{m_a}P_d}
\end{align}

Where $C_x$ is scaling coefficient obtained from analytical modelling, which makes equations linear. Generally it has 95\% prediction band but as was said earlier this linearity is what gives problems to current thrusters designs (high mass, same power density, average performance). The original dataset is
\begin{table}[hb!]
    \centering
    \begin{tabular}{l|l|l|l|l|l|l|l|l}
    	\toprule
    	\midrule
        Thruster & Power, W & $U_{d}$, V & d, mm & h, mm & L, mm & $m_{\dot{a}}$, \text{mg/s} & T, N & $I_{spa}$, s \\ 
        SPT-20 & 52.4 & 180 & 15.0 & 5.0 & 32.0 & 0.47 & 3.9 & 839 \\ 
        SPT-25 & 134 & 180 & 20.0 & 5.0 & 10 & 0.59 & 5.5 & 948 \\
        Music-si & 140 & 288 & 18 & 2 & 6.5 & 0.44 & 4.2 & 850 \\ \midrule
        HET-100 & 174 & 300 & 23.5 & 5.5 & 14.5 & 0.50 & 6.8 & 1386 \\ 
        KHT-40 & 187 & 325 & 31.0 & 9.0 & 25.5 & 0.69 & 10.3 & 1519 \\ 
        KHT-50  & 193 & 250 & 42.0 & 8.0 & 25.0 & 0.88 & 11.6 & 1339 \\ 
        HEPS-200 & 195 & 250 & 42.5 & 8.5 & 25.0 & 0.88 & 11.2 & 1300 \\ 
        BHT-200 & 200 & 250 & 21.0 & 5.6 & 11.2 & 0.94 & 12.8 & 1390 \\ 
        KM-32 & 215 & 250 & 32.0 & 7.0 & 16.0 & 1.00 & 12.2 & 1244 \\  \midrule
        ...\\ \midrule
        HEPS-500 & 482 & 300 & 49.5 & 15.5 & 25.0 & 1.67 & 25.9 & 1587 \\ \midrule
        UAH-78AM & 520 & 260 & 78.0 & 20 & 40 & 2 & 30 & 1450 \\ 
        BHT-600 & 615 & 300 & 56.0 & 16.0 & 32 & 2.60 & 39.1 & 1530 \\ 
        SPT-70 & 660 & 300 & 56.0 & 14.0 & 25.0 & 2.56 & 40.0 & 1593 \\ 
        MaSMi60 & 700 & 250 & 60 & 9.42 & 19 & 2.56 & 30 & 1300 \\ \midrule
        MaSMiDm & 1000 & 500 & 67 & 10.5 & 21 & 3 & 53 & 1940 \\ 
        SPT-100  & 1350 & 300 & 85.0 & 15.0 & 25.0 & 5.14 & 81.6 & 1540 \\ \midrule
        \bottomrule
    \end{tabular}
\end{table}
Hosting only 24 entries in total. The references are as follows\cite{beal_plasma_2004}\cite{belikov_high-performance_2001}\cite{kronhaus_discharge_2013}\cite{misuri_het_2008}\cite{lee_scaling_2019}

In the next section the used neural networks architectures will be discussed.

\section{Data driven HET designs}

Neural networks are a type of machine learning algorithm that is often used in the field of artificial intelligence. They are mathematical models that can be trained to recognize patterns within large datasets. The architecture of GAN's generator was already shown. In this section we will focus on fully connected networks, which are most popular for type for these tasks. HETFit code leverages dynamic architecture generation of these FcNN's which is done via meta learning algorithm Tree-structured Parzen Estimator for every data input user selects. This code uses state-of-art implementation made by OPTUNA. The dynamically suggested architecture has 2 to 6 layers from 4 to 128 nodes on each with SELU, Tanh or ReLU activations and most optimal optimizer. 
The code user interface is as follows:
1. Specify working environment
2. Load or generate data
3. Tune the architecture
4. Train and get robust scaling models
\begin{figure}[t!]
\centering
\includegraphics[width=.25\textwidth]{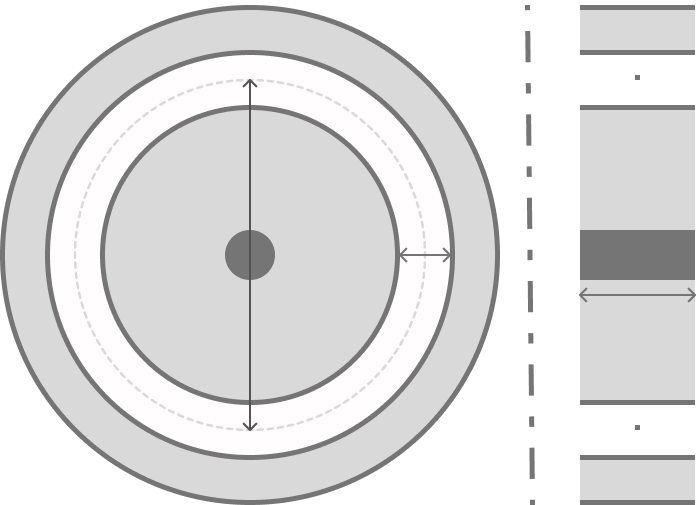}
\caption{Hall effect thruster geometry}
\end{figure}

\subsection{FNN}
All of Fully connected neural networks are implemented in PyTorch as it the most powerful ML/AI library for experiments. When the network architecture is generated, all of networks have similar training loops as they use gradient descend algorithm :
Loss function: \begin{equation}
    L(w, b) \equiv \frac{1}{2 n} \sum_x\|y(x)-a\|^2
\end{equation}
This one is mean square error (MSE) error function most commonly used in FNNs. Next we iterate while updating weights for a number of specified epochs this way.
Loop for number of epochs:

- Get predictions: $\hat{y}$

- Compute loss: $\mathscr{L}(w, b)$

- Make backward pass

- Update optimizer

It can be mentioned that dataset of electric propulsion is extremely complex due to large deviations in data. Thanks to adavnces in data science and ML it is possible to work with it.

This way we assembled dataset on our ROI domain of $P$<1000 $W$ input power and 200-500 $V$ range. Sadly one of limitations of such model is disability to go beyond actual database limit while not sacrificing performance and accuracy.

\subsection{Physics Informed Neural Networks}
For working with unscaled data PINN's were introduced, they are using equations 2-7 to generate $C_x$ coefficients. Yes, it was said earlier that this method lacks ability to generate better performing HETs, but as we have generated larger dataset on same domain as \citet{lee_scaling_2019} it is important to control that our dataset is still the same quality as original. Using above mentioned PINN's it was possible to fit coefficients and they showed only slight divergence in values of few \% which is acceptable.

\subsection{ML approach notes}
We already have discussed how HETFit code works and results it can generate, the overiew is going to be given in next section. But here i want to warn that this work is highly experimental and you should always take ML approaches with a grain of salt, as some plasma discharge physics in HET is yet to be understood, data driven way may have some errors in predictions on specific bands.
Few notes on design tool I have developed in this work: it is meant to be used by people with little to no experience in ML field but those who wants to quickly analyze their designs or create baseline one for simulations. One can even use this tool for general tabular data as it has mostly no limits whatsoever to input data.

\subsection{Two input variables prediction}
One of main characteristics for any type of thruster is efficiency, in this work I researched dependency of multiple input values to $\eta_t$. Results are as follows in form of predicted matrix visualisations. Figure 3 takes into account all previous ones in the same time, once again it would be way harder to do without ML.
\begin{figure}[h!]
    \begin{minipage}[c]{0.5\linewidth} 
        \centering
        \includegraphics[width=\textwidth]{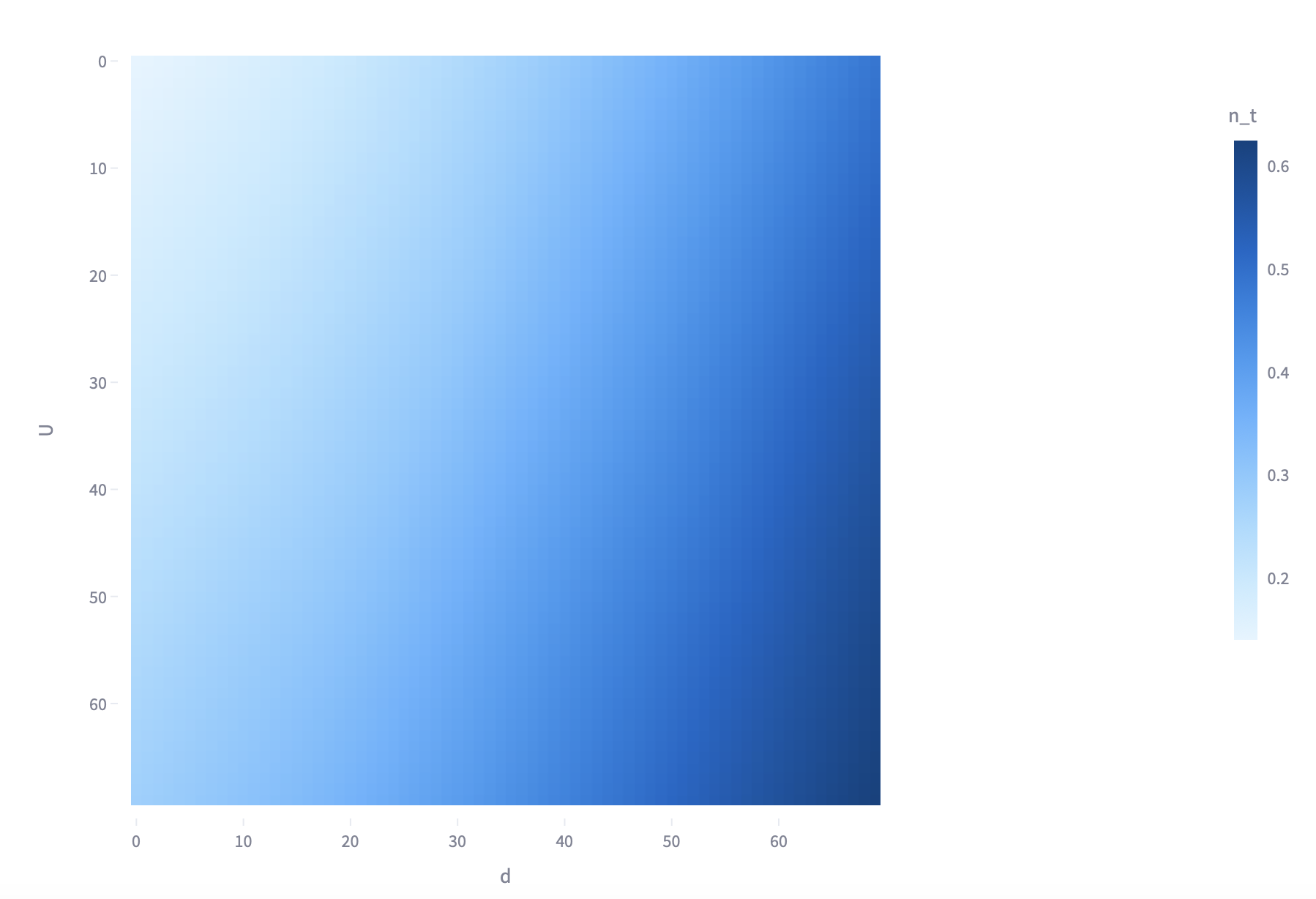}
        \caption{$U,d \to \eta_t$ predictions}
    \end{minipage}\hfill
    \begin{minipage}[c]{0.5\linewidth} 
        \centering
        \includegraphics[width=\textwidth]{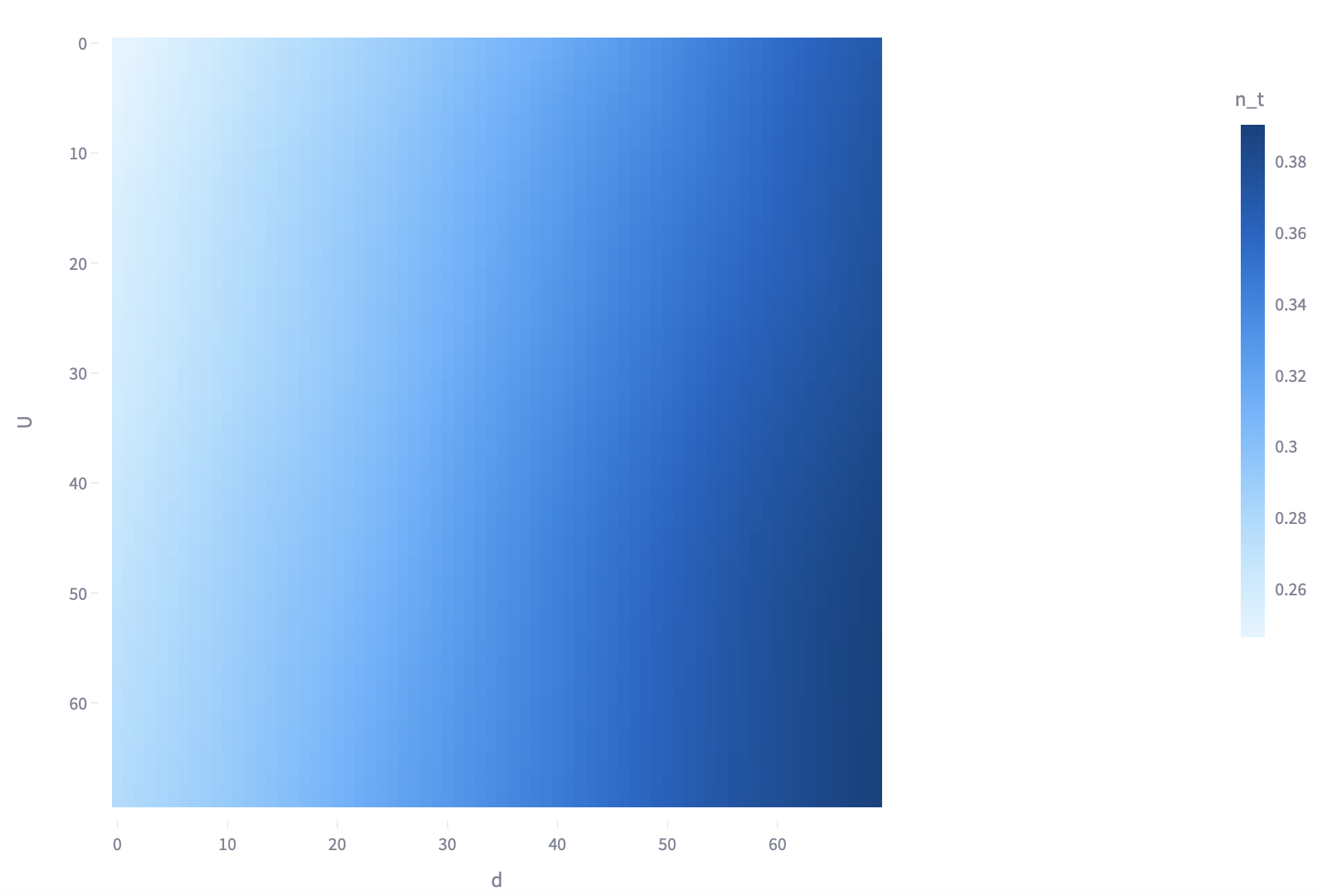}
        \caption{$U,d \to \eta_t$ predictions}
    \end{minipage}
    \begin{minipage}[c]{0.5\linewidth} 
        \centering
        \includegraphics[width=\textwidth]{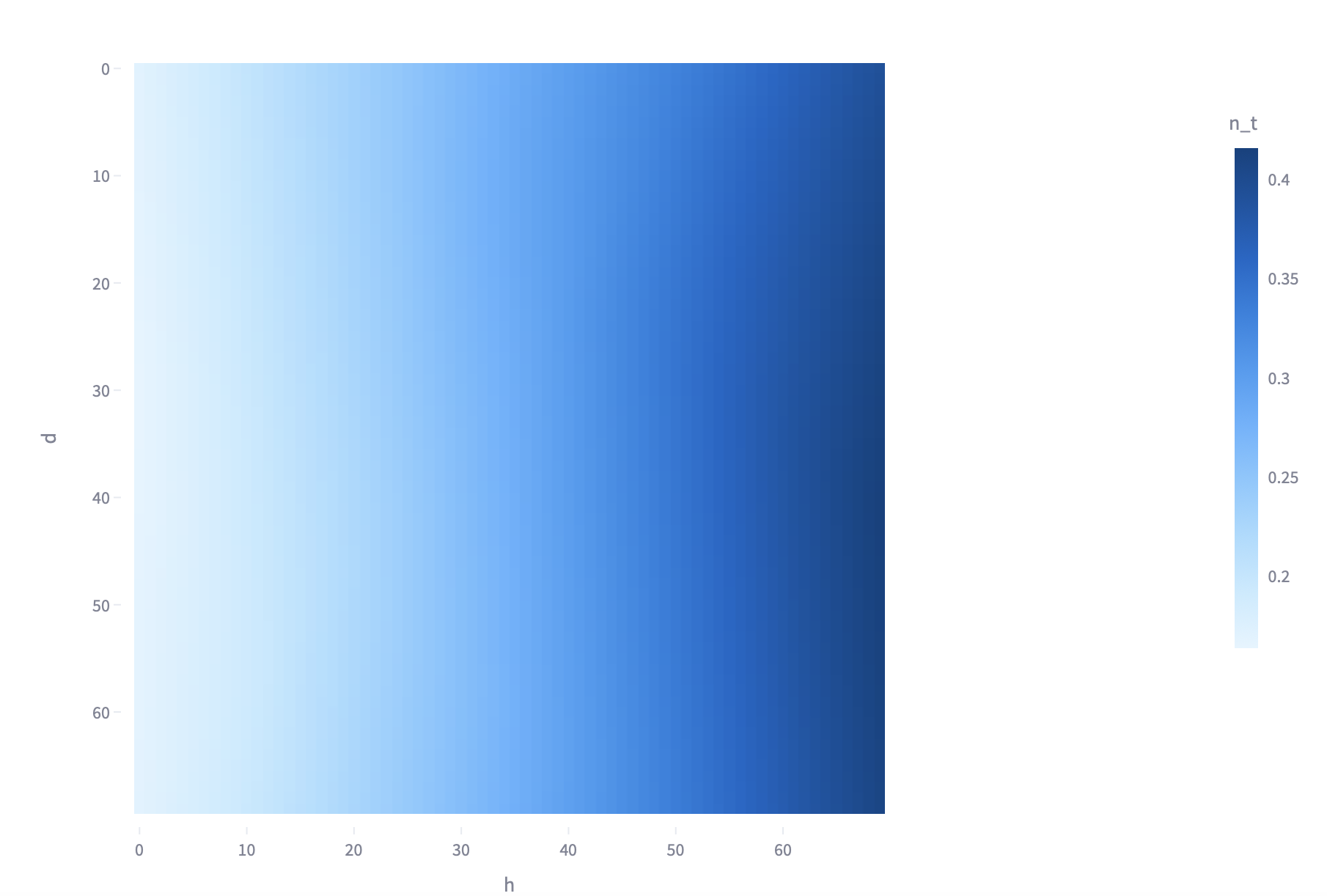}
        \caption{$d,h \to \eta_t$ predictions}
    \end{minipage}\hfill
    \begin{minipage}[c]{0.5\linewidth} 
        \centering
        \includegraphics[width=\textwidth]{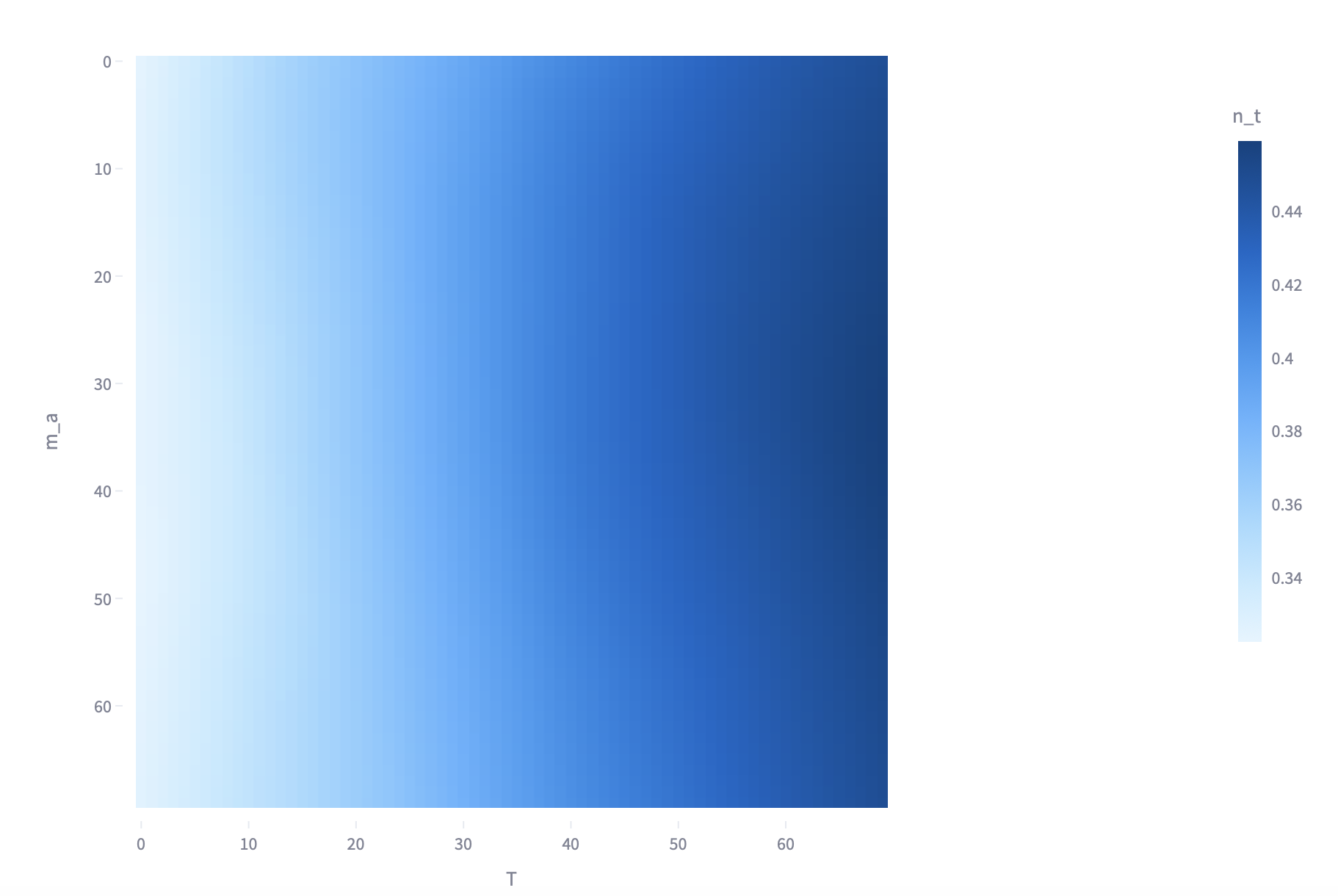}
        \caption{$m_{\dot{a}},T \to \eta_t$ predictions}
    \end{minipage}%
\end{figure}
\begin{figure}[h!]
\centering
\includegraphics[width=.5\textwidth]{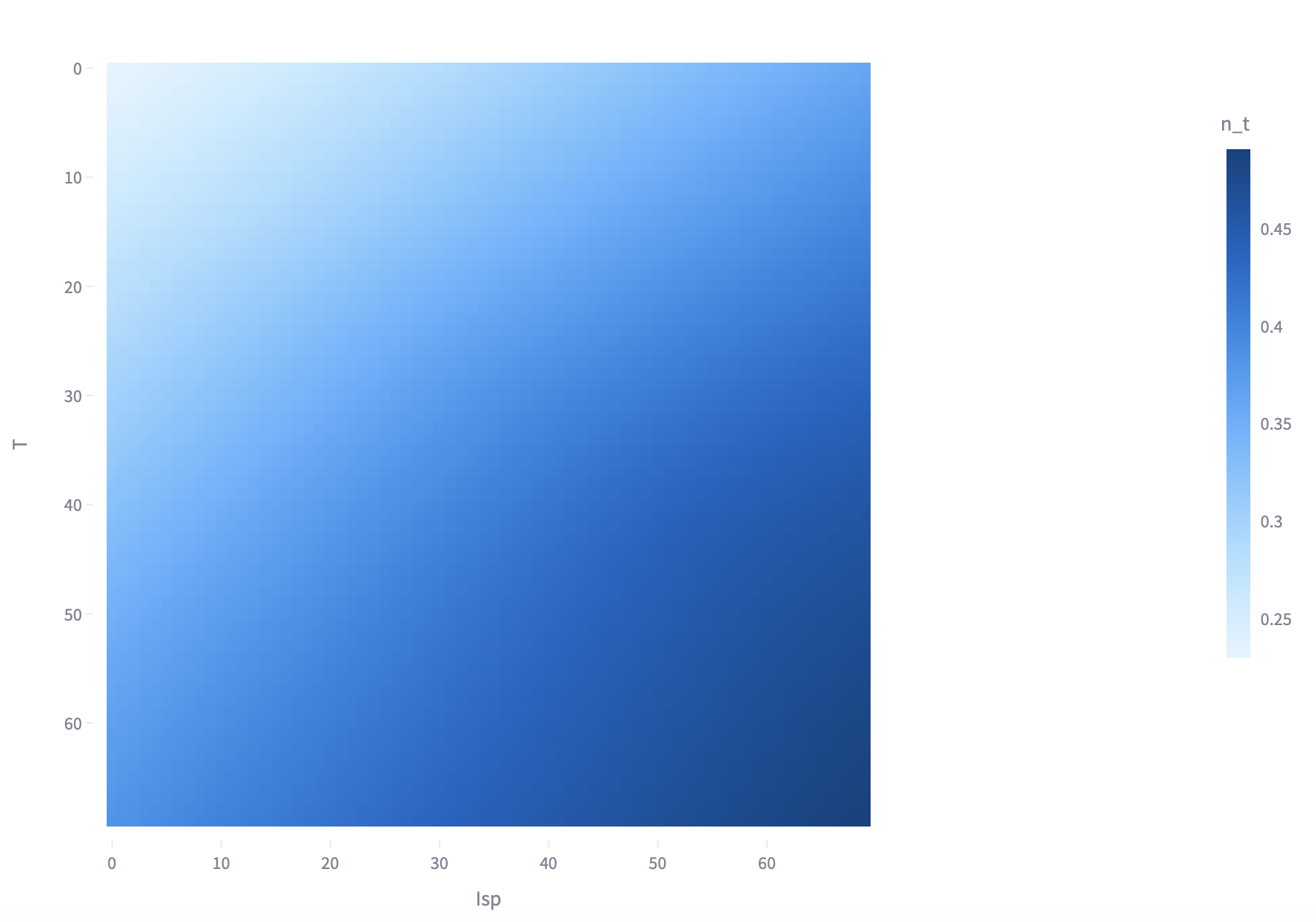}
\caption{$I_{sp},T \to \eta_t$ predictions}
\end{figure}

\section{Results discussion}

Let's compare predictions of semi empirical approach\cite{lee_scaling_2019}, approach in paper\cite{plyashkov_scaling_2022}, and finally ours. Worth to mention that current approach is easiest to redesign from scratch.

\subsection{NN architecture generation algorithm}
As with 50 iterations, previously discussed meta learning model is able to create architecture with score of 0.9+ in matter of seconds. HETFit allows logging into neptune.ai environment for full control over simulations. Example trail run looks like that.
\begin{figure}[h!]
\centering
\includegraphics[width=.8\textwidth]{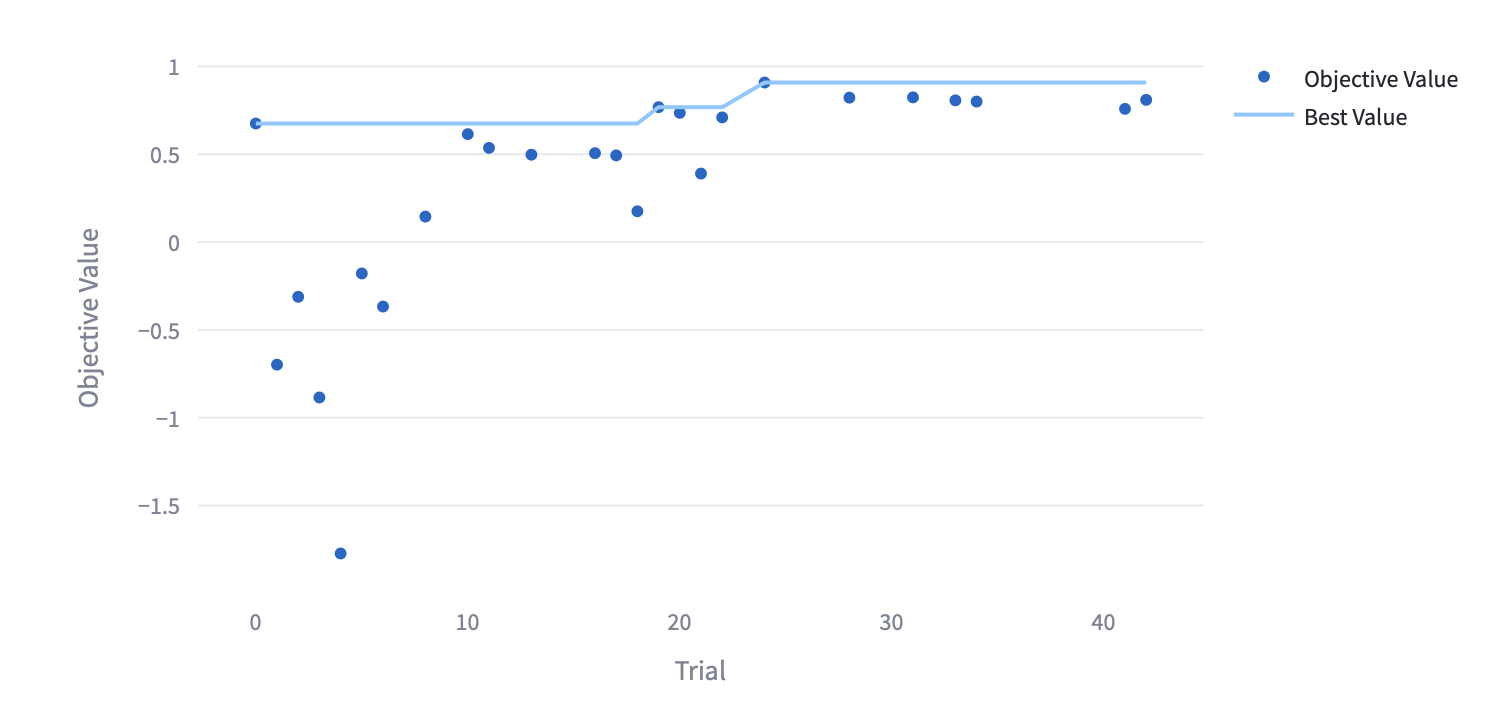}
\caption{TPE algorithm architecture optimization}
\end{figure}

\subsection{Power density and magnetic flux dependence}

Neither of the models currently support taking magnetic flux in account besides general physics relations, but we are planning on updating the model in next follow up paper. For now $\vec{B}$ relation to power remains unresolved to ML approach but the magnetic field distribution on z axis is computable and looks like that for magnetically shielded thrusters:
\begin{figure}[h!]
\centering
\includegraphics[width=.8\textwidth]{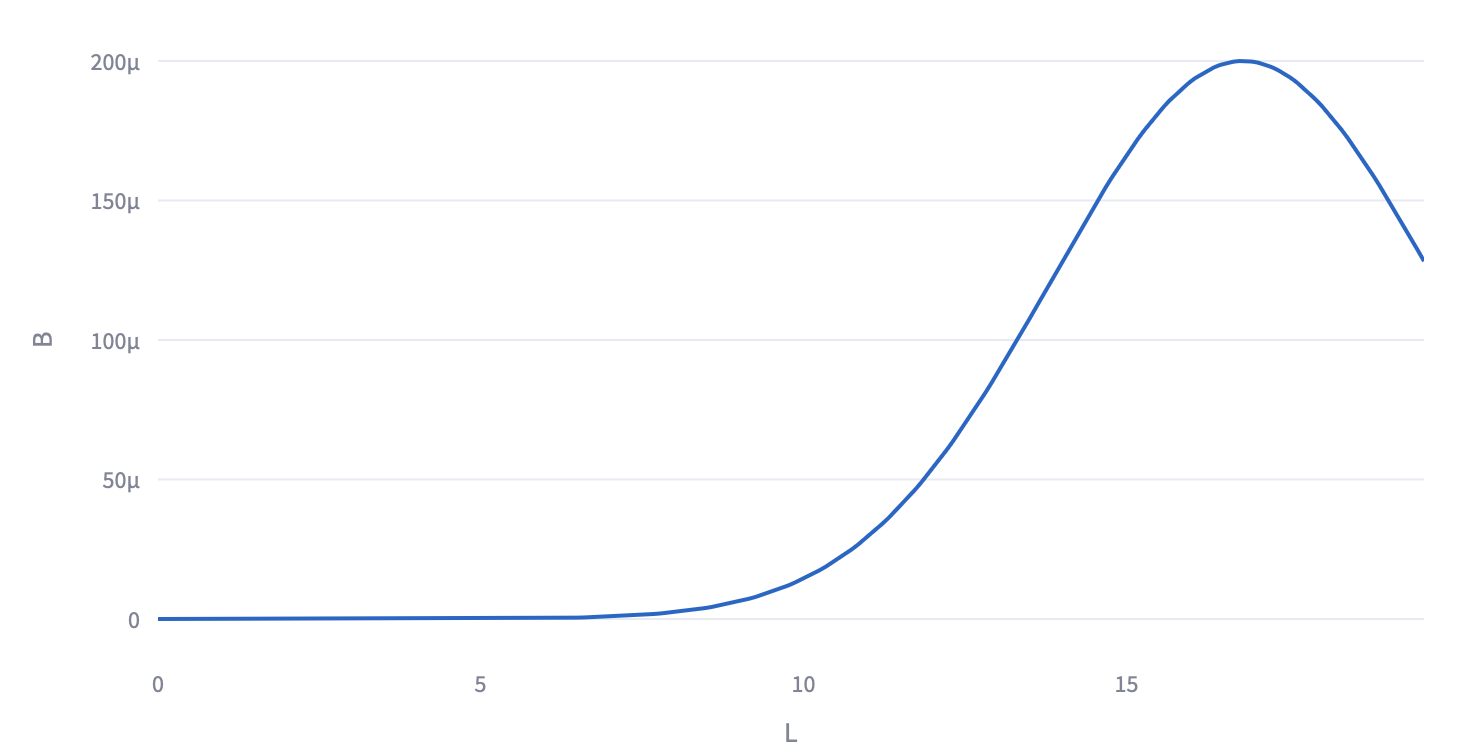}
\caption{dB/dz distribution}
\end{figure}

\subsection{Dependency of T on d,P}

Following graph is describing Thrust as function of channel diameter and width, where hue map is thrust. It is well known dependency and it has few around 95\% prediction band \citep{lee_scaling_2019}

\begin{figure}[h!]
\centering
\includegraphics[width=0.6\textwidth]{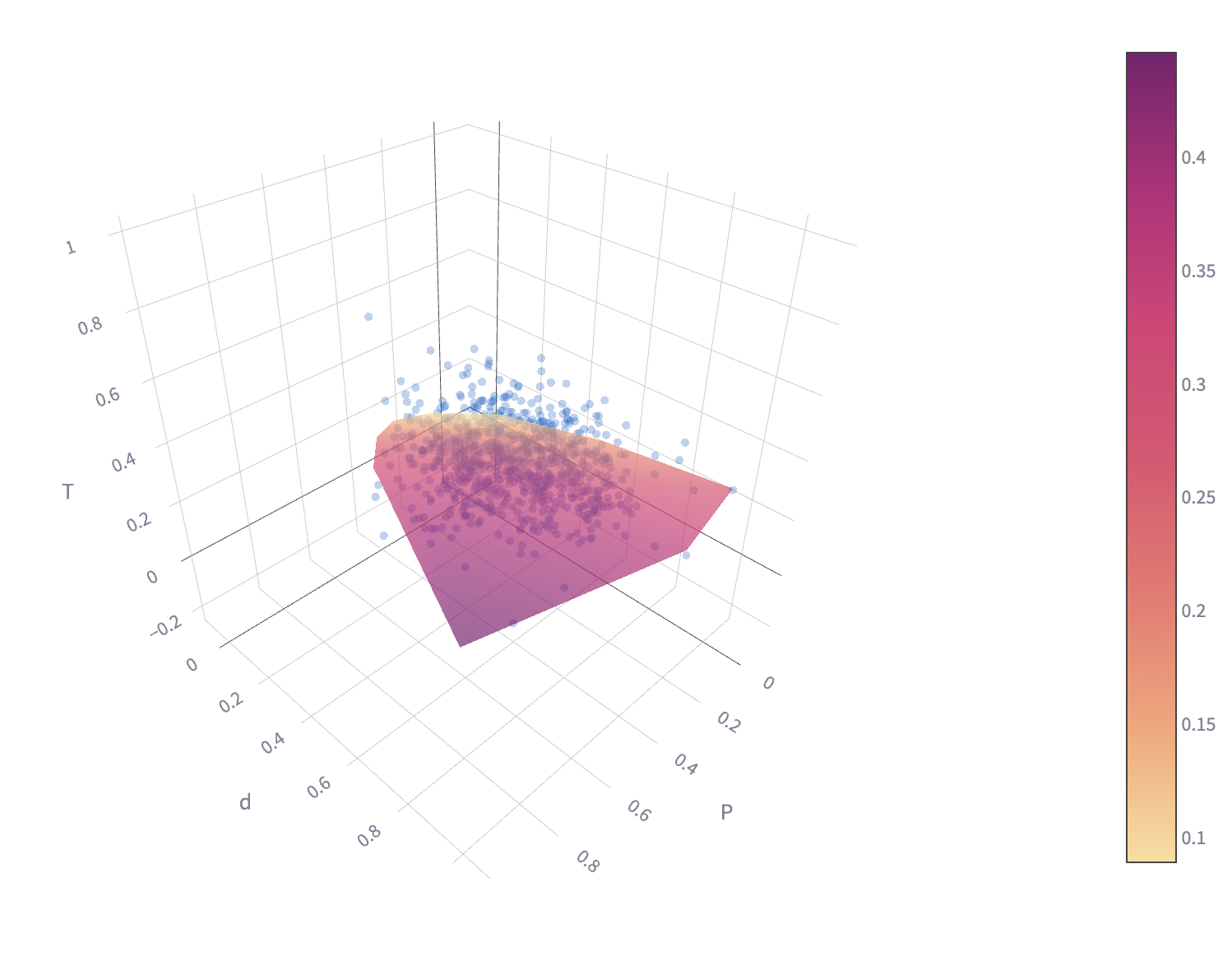}
\caption{Dependency of T on d,P}
\end{figure}

\subsection{Dependency of T on P,U}

\begin{figure}[h!]
\centering
\includegraphics[width=.6\textwidth]{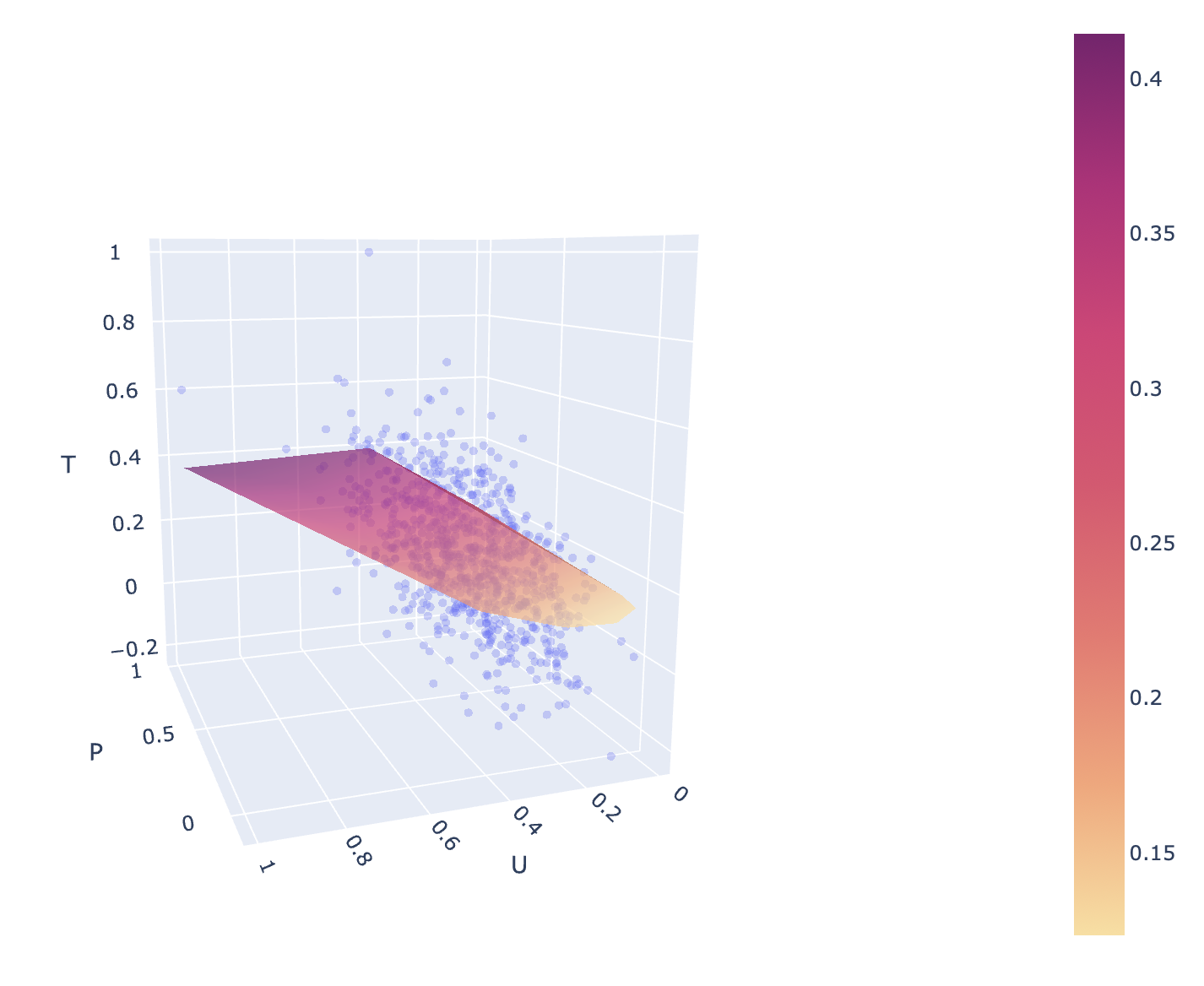}
\caption{Dependency of T on P,U}
\end{figure}

\subsection{Dependency of T on $m_a$,P}

Compared to\citep{shagayda_hall-thruster_2013} The model accounts for more parameters than linear relation. So such method proves to be more precise on specified domain than semi empirical linear relations.

\begin{figure}[h!]
\centering
\includegraphics[width=.6\textwidth]{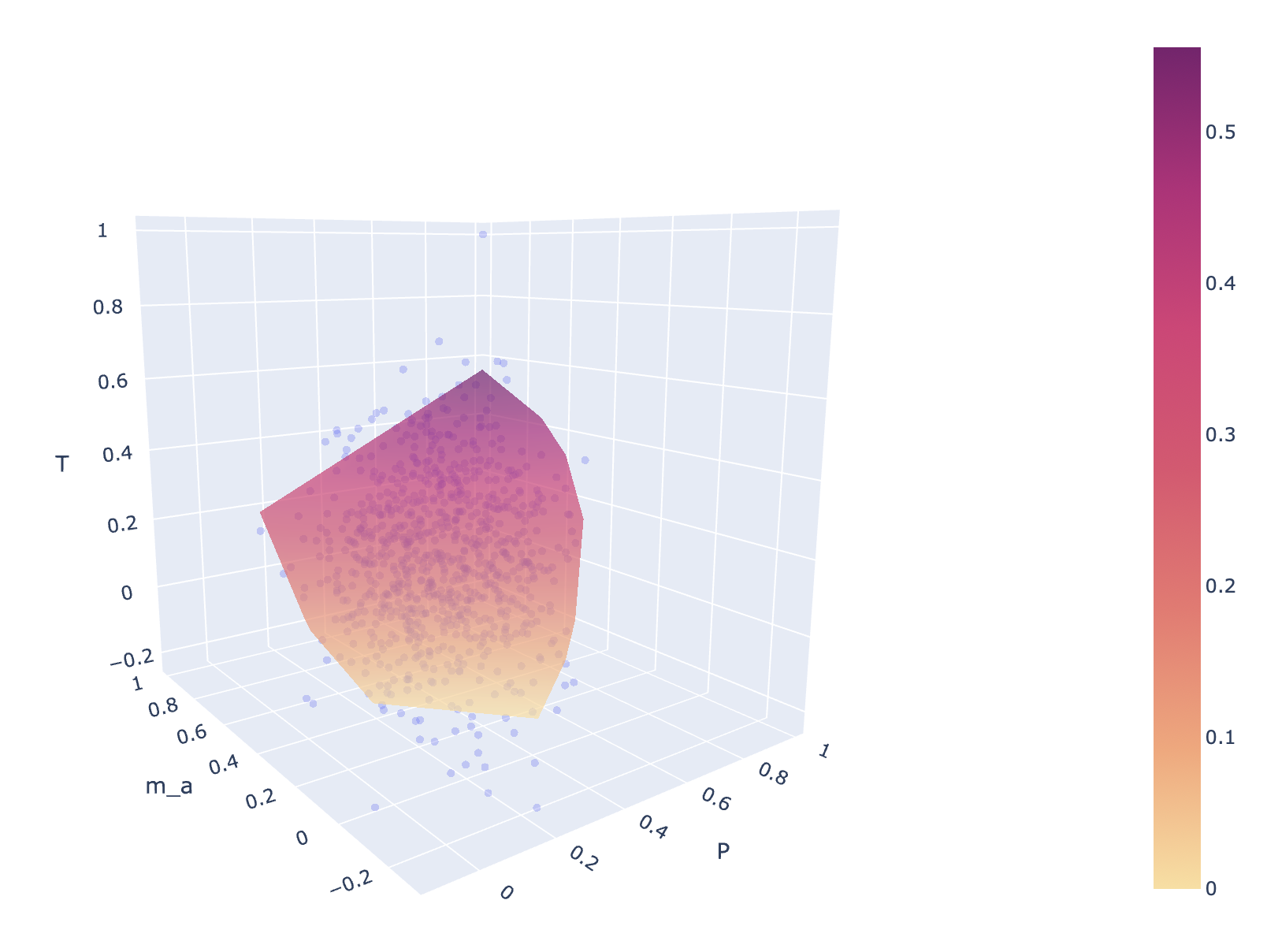}
\caption{Dependency of T on $m_a$,P}
\end{figure}

\subsection{Dependency of $I_{sp}$ on d,h}

\begin{figure}[h!]
\centering
\includegraphics[width=.6\textwidth]{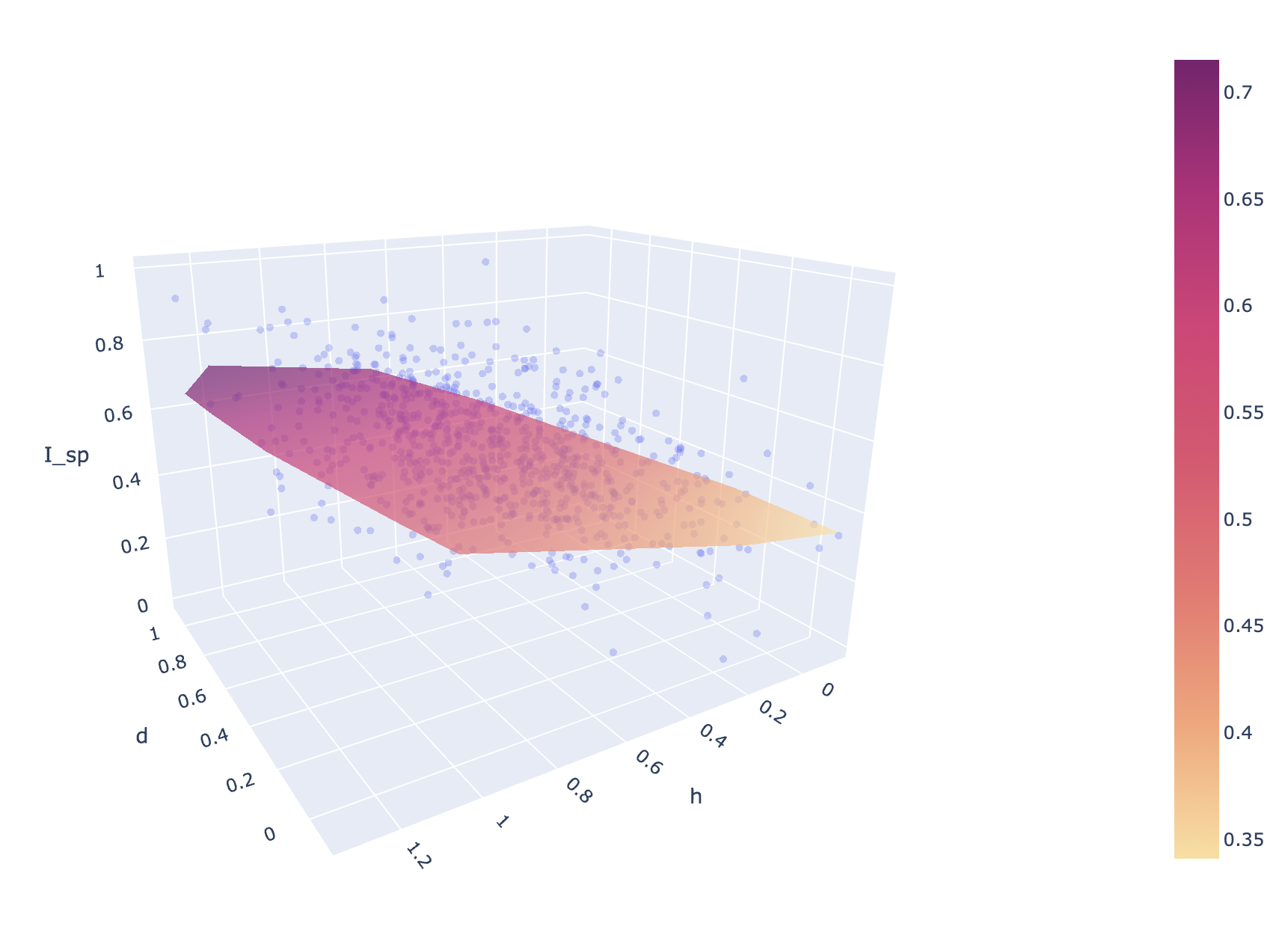}
\caption{Dependency of $I_{sp}$ on d,h}
\end{figure}

We generated many models so far, but using ML we can make single model for all of the parameters at the same time, so these graphs tend to be 3d projection of such model inference.

\subsection{Use of pretrained model in additive manufacturing of hall effect thruster channels}

The above mentioned model was used to predict geometry of channel, next the simulation was conducted on this channel. Second one for comparison was calculated via usual scaling laws. The initial conditions for both are:
\begin{table}[h!]
    \centering
    \begin{tabular}{l|l}
    	\toprule
    	\midrule
    	Initial condition & Value \\
        \midrule
        $n_{e,0}$ & $1e13 [m^{-3}]$\\
        $\epsilon_0$ & 4 [V]\\
        V & 300 [V]\\
        T & 293.15 [K]\\
        $P_{abs}$ & $0.5$ [torr]\\
        $\mu_e N_n$ & $1e25$ $[1/(V\cdot m \cdot s)]$\\
        dt & 1e-8 [s]\\
        Body & Ar\\
        \bottomrule
    \end{tabular}
\end{table}

Outcomes are so that ML geometry results in higher density generation of ions which leads to more efficient thrust generation. HETFit code suggests HET parameters by lower estimate to compensate for not included variables in model of HET. This is experimentally proven to be efficient estimate since SEM predictions of thrust are always higher than real performance.
\citet{lee_scaling_2019}

\begin{figure}[hpt!]
\centering
\includegraphics[width=.6\textwidth]{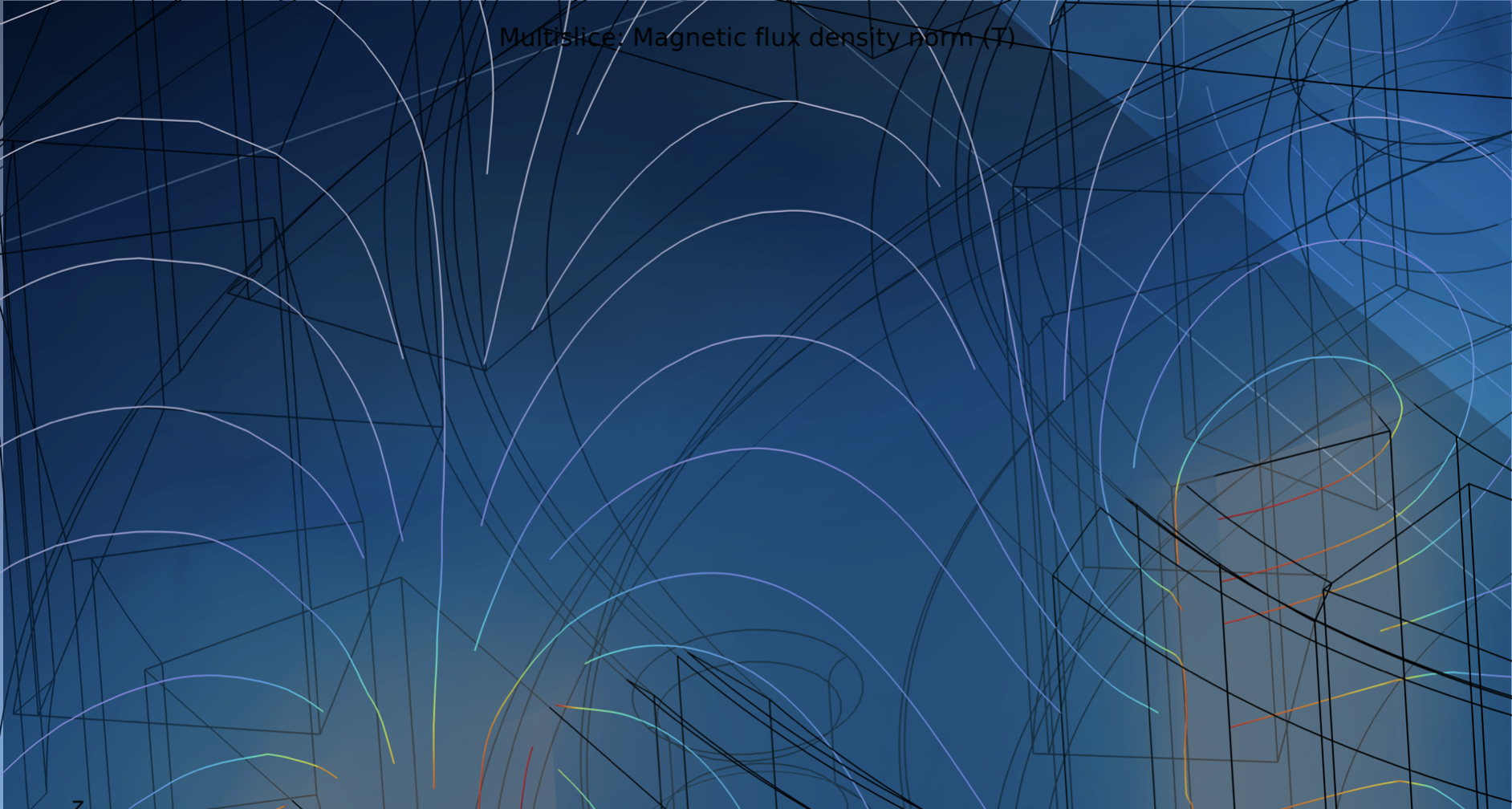}
\caption{Magnetic flux density distribution, magnetic shielding configuration}
\end{figure}

\subsection{Code description}

Main concepts:
- Each observational/design session is called an environment, for now it can be either RCI or SCI (Real or scaled interface) 

- Most of the run parameters are specified on this object initialization, including generation of new samples via GAN

- Built-in feature generation (log10 Power, efficiency, $\vec{B}$, etc.)

- Top feature selection for each case. (Boruta algorithm)

- Compilation of environment with model of choice, can be any torch model or sklearn one

- Training

- Plot, inference, save, export to jit/onnx, measure performance 

\subsection{COMSOL HET simulations}
The simulations were conducted in COMSOL in plasma physics interface which gives the ability to accurately compute Electron densities, temperatures, energy distribution functions from initial conditions and geometry. Here is comparison of both channels.
\clearpage

\begin{figure}
\centering
\includegraphics[width=.8\textwidth]{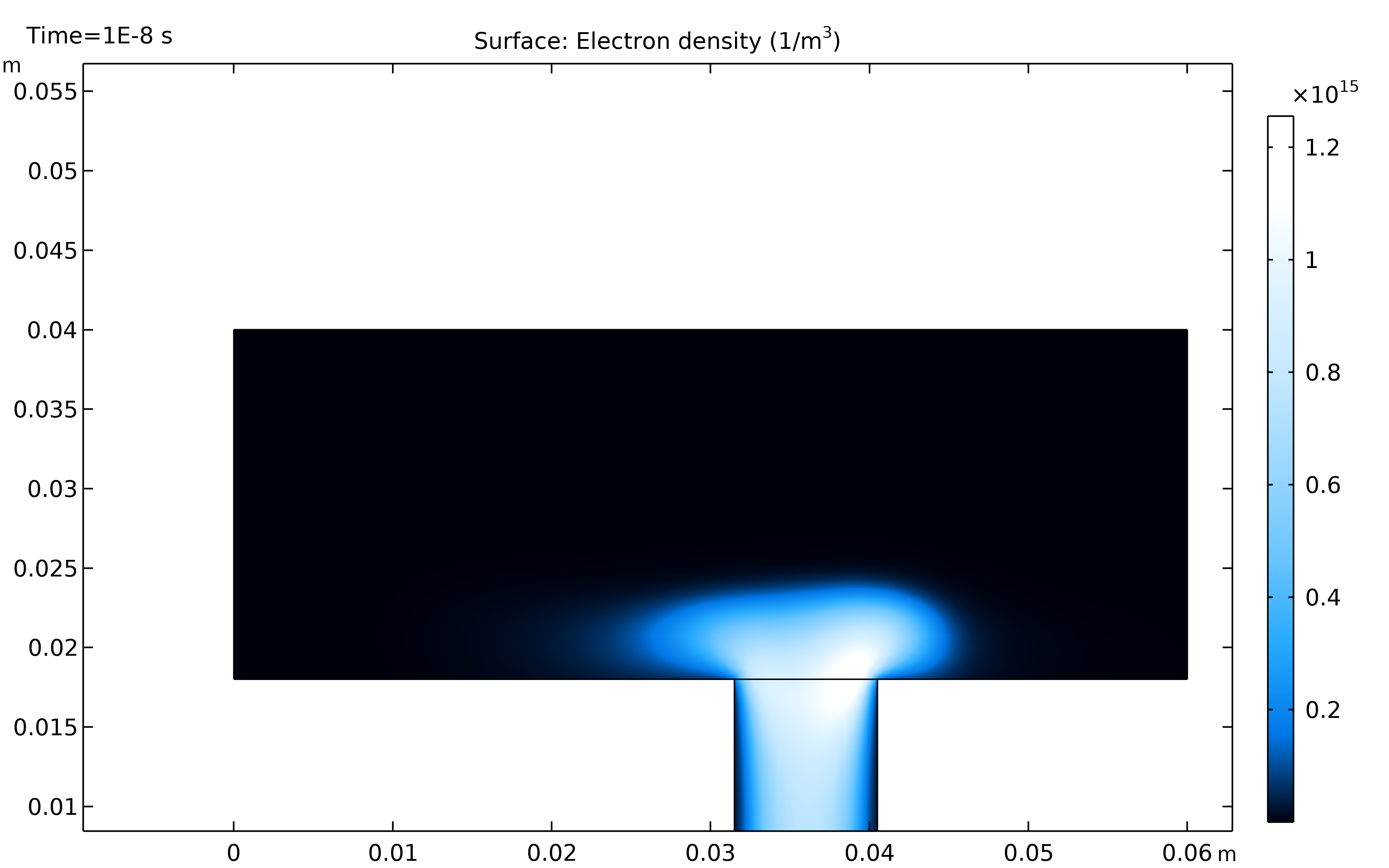}
\caption{Electron density with linear SEM geometry}
\end{figure}

\begin{figure}
\centering
\includegraphics[width=.8\textwidth]{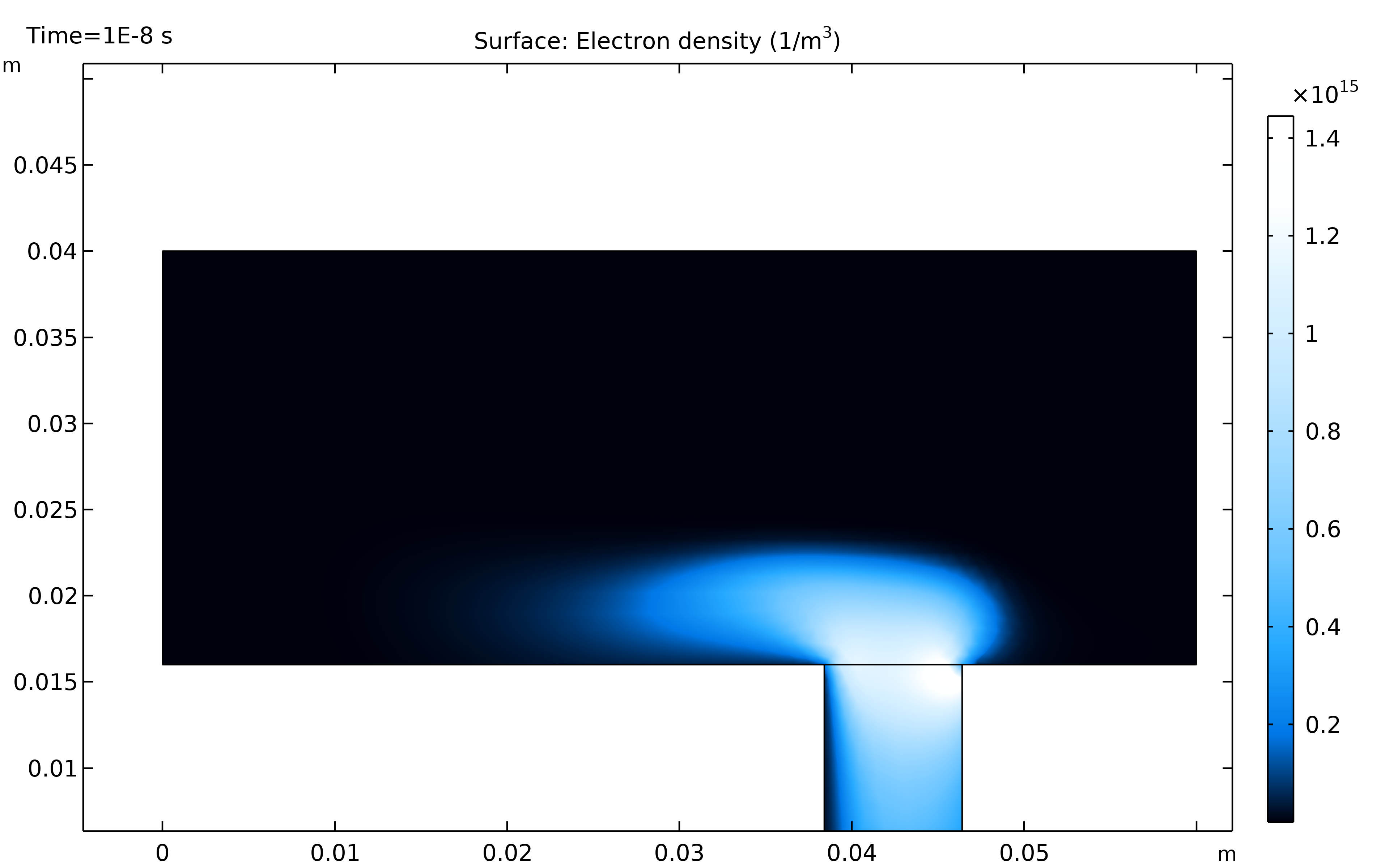}
\caption{Electron density with HETfit geometry}
\end{figure}
\clearpage

\section{Conclusion}
In conclusion the another model of scaling laws was made and presented. HETFit code is open source and free to be used by anyone. Additively manufactured channel was printed to prove it's manufactureability. Hopefully this work will help developing more modern scaling relations as current ones are far from perfect.

Method in this paper and firstly used in \citet*{plyashkov_scaling_2022} has advantages over SEM one in: ability to preidct performance more precisely on given domain, account for experimental data. I believe with more input data the ML method of deisgning thrusters would be more widely used.

The code in this work could be used with other tabular experimental data since most of cases and tasks tend to be the same: feature selection and model optimization.

\begin{figure}
\centering
\includegraphics[width=.7\textwidth]{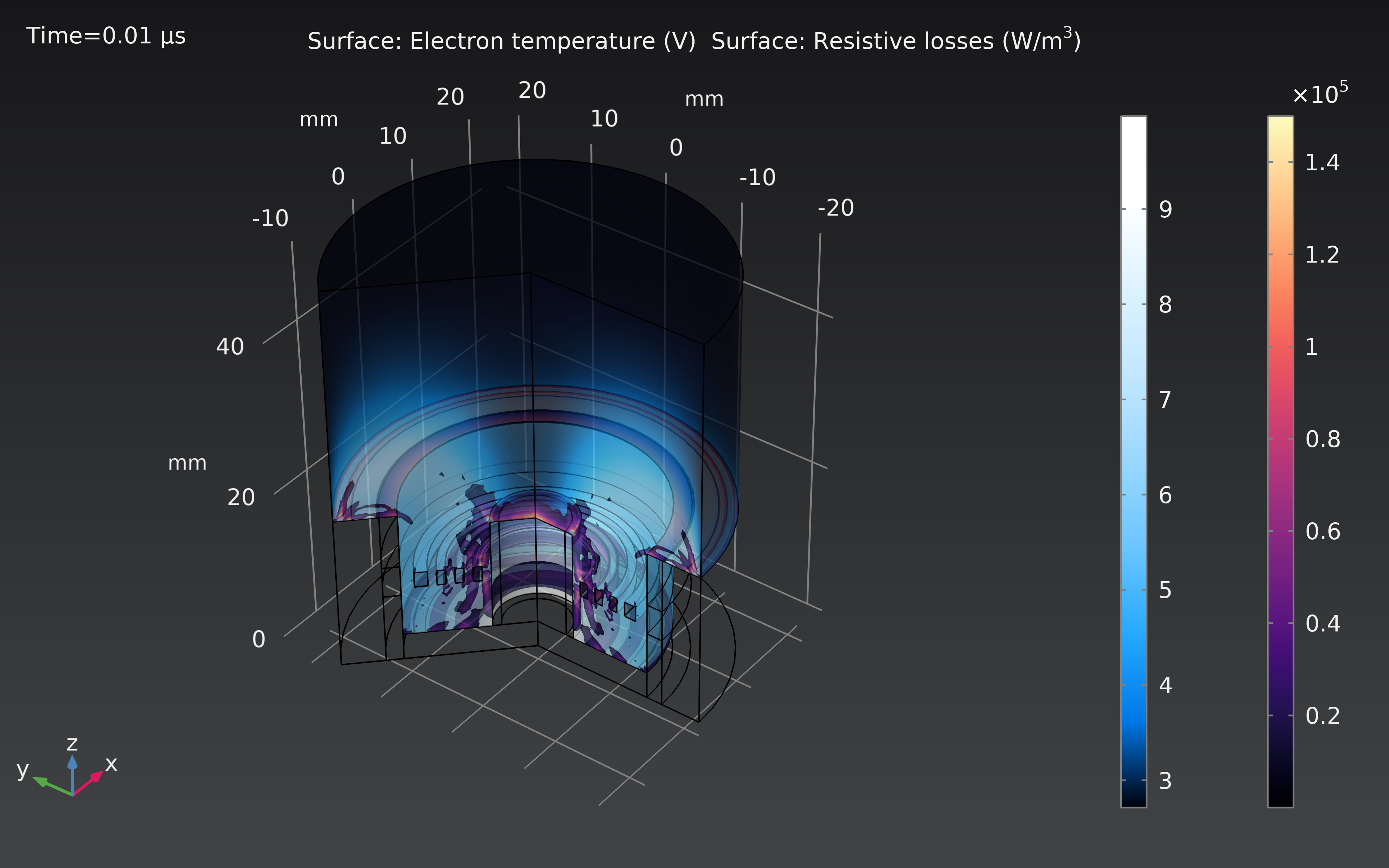}
\caption{COMSOL simulation of designed thruster start up}
\end{figure}
\begin{figure}
\centering
\includegraphics[width=.35\textwidth]{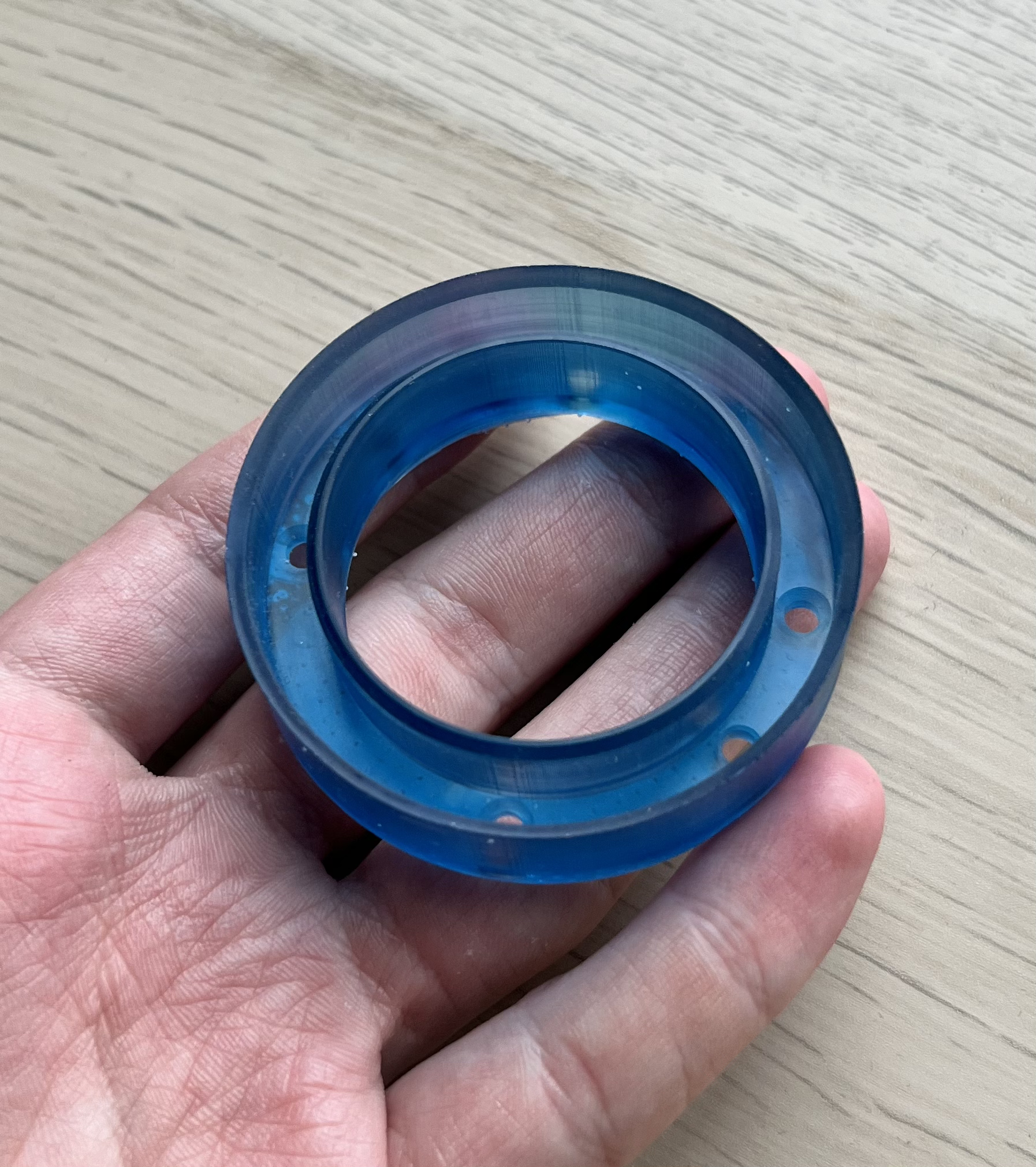}
\caption{Manufactured channel}
\end{figure}

\clearpage
\bibliography{sample}

\end{document}